\documentclass[10pt,twocolumn,letterpaper]{article}

\usepackage{iccv}
\usepackage{times}
\usepackage{epsfig}
\usepackage{graphicx}
\usepackage{amsmath}
\usepackage{amssymb}

\usepackage{pifont}
\usepackage{booktabs}
\usepackage{multirow}
\usepackage{capt-of}
\usepackage{makecell}
\usepackage{colortbl}
\usepackage{dsfont}
\usepackage{comment}
\usepackage[dvipsnames]{xcolor}

\usepackage[accsupp]{axessibility}  %

\usepackage{tocloft}

\makeatletter
\renewcommand{\numberline}[1]{%
  \@cftbsnum #1\@cftasnum\hspace*{1em}\@cftasnumb%
}
\makeatother

\usepackage[pagebackref=true,breaklinks=true,colorlinks,bookmarks=false]{hyperref}

\usepackage[capitalize]{cleveref}
\crefname{section}{Sec.}{Secs.}
\Crefname{section}{Section}{Sections}
\Crefname{table}{Table}{Tables}
\crefname{table}{Tab.}{Tabs.}

\renewcommand{\paragraph}[1]{\vspace{1.25mm}\noindent\textbf{#1}}

\definecolor{baselinecolor}{gray}{.9}
\definecolor{darkgreen}{rgb}{0.13, 0.55, 0.13}

\let\originalleft\left
\let\originalright\right
\renewcommand{\left}{\mathopen{}\mathclose\bgroup\originalleft}
\renewcommand{\right}{\aftergroup\egroup\originalright}

\iccvfinalcopy %

\def\iccvPaperID{5097} %

\begin{document}

\title{Knowledge Adaptation Network for Few-Shot Class-Incremental Learning}

\author{%
  Ye Wang\textsuperscript{$1$} \qquad
  Yaxiong Wang\textsuperscript{$2$} \qquad
  Guoshuai Zhao\textsuperscript{$1$} \qquad
  Xueming Qian\textsuperscript{$1$} \\
  \textsuperscript{$1$}Xi'an Jiaotong University \quad \\
  \textsuperscript{$2$}Hefei University of Technology \quad \\
  \tt\small \{xjtu2wangye@stu,wangyx15@stu,guoshuai.zhao@, qianxm@mail\}.xjtu.edu.cn \\
}



\maketitle


\begin{abstract}
Few-shot class-incremental learning (FSCIL) aims to incrementally recognize new classes using a few samples while maintaining the performance on previously learned classes. One of the effective methods to solve this challenge is to construct prototypical evolution classifiers. Despite the advancement achieved by most existing methods, the classifier weights are simply initialized using mean features. Because representations for new classes are weak and biased, we argue such a strategy is suboptimal. In this paper, we tackle this issue from two aspects. Firstly, thanks to the development of foundation models, we employ a foundation model, the CLIP, as the network pedestal to provide a general representation for each class. Secondly, to generate a more reliable and comprehensive instance representation, we propose a Knowledge Adapter (KA) module that summarizes the data-specific knowledge from training data and fuses it into the general representation. Additionally, to tune the knowledge learned from the base classes to the upcoming classes, we propose a mechanism of Incremental Pseudo Episode Learning (IPEL) by simulating the actual FSCIL. Taken together, our proposed method, dubbed as Knowledge Adaptation Network (KANet), achieves competitive performance on a wide range of datasets, including CIFAR100, CUB200, and ImageNet-R. 
\end{abstract}

\section{Introduction}
\label{sec:intro}

Human beings are naturally endowed with the ability to continuously learn new knowledge without forgetting old knowledge. However, it's a nontrivial task for most existing deep models\cite{ResNet,PINTELAS2023103687,NAM2023103684}. To empower deep models with incremental learning ability, many researchers engage in the research known as Class-Incremental Learning (CIL) and propose many elegant and effective methods~\cite{wang2022dualprompt, JI2021107589, WANG2022109052}. As one of the critical components for achieving such success, sufficient training samples are accessible for most existing CIL methods to learn new classes. In some scenarios, \eg, foreign object recognition on railway tracks, the number of available training data is limited, making these methods often suffer from the overfitting problem. Regarding this, the task of few-shot class-incremental learning (FSCIL)~\cite{Tao_2020_CVPR} is designed to incrementally learn new classes using a few samples while not forgetting previously learned classes.

Due to the practical and challenging nature of FSCIL, the research interest of many scholars is ignited for this task. The mainstream FSCIL works~\cite{Zhu_2021_CVPR,10168925,wang2023fewshot} solve this task by model decoupling strategy, which means the encoder is frozen in incremental learning sessions and only the classifier weights or features are updated with various modules. Despite advancements achieved by these methods, most are built on the model trained from scratch. Under the context of FSCIL, a fatal shortcoming of such a learning framework is that new data representations are essentially weak, resulting in weak performance in incremental sessions as evidenced in these methods. For this issue, the remarkable success in representation achieved by  CLIP~\cite{radford2021learning} leads us to believe that CLIP is a promising FSCIL solver. Nevertheless, due to the intrinsic challenges of FSCIL and lack of task-specific knowledge, adapting the CLIP to this task is nontrivial. Concretely, to adapt the CLIP to FSCIL well, we must address a major challenge, \ie how to efficiently tune the CLIP to match the task-specific context of FSCIL using limited samples.

\begin{figure*}
    \centering
    \includegraphics[width=1.8\columnwidth]{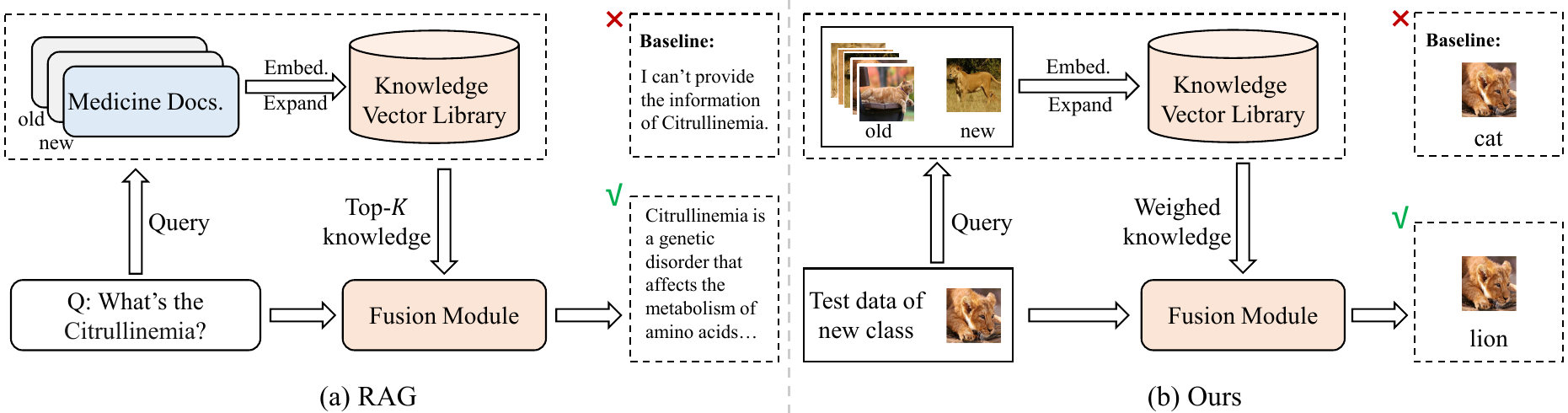}
    \caption{Illustration of our motivation. Our proposed method borrows the treasure from Retrieval Augmented Generation (RAG) technique to summarize the data-specific knowledge into a Knowledge Vector Library (KVL) and use them to refine the model's output.}
    \label{fig:motiv}
\end{figure*}

To tackle this challenge, we propose a Knowledge Adapter (KA) module (see Section\textcolor{red}{~\ref{sec:KA}}). Concretely, the KA learns the task-specific knowledge of FSCIL with adaptation parameters and utilizes a query-based knowledge fusion (QKF) mechanism to integrate the task-specific knowledge into the CLIP using adaptation parameters. More precisely, our proposed method defines the knowledge as refining the representation to better fit incremental sessions. However, the remaining problem is how to refine the representation. In fact, when applying general-purpose LLMs to specific scenarios, such as the medical knowledge Q\&A (Figure\textcolor{red}{~\ref{fig:motiv}}(a)), the lack of expertise knowledge often leads to suboptimal answers. To address this issue, one commonly used approach is the Retrieval Augmented Generation (RAG) technique~\cite{wu2024continual}. Concretely, RAG stores data-specific knowledge in a Knowledge Vector Library (KVL). When processing professional questions, RAG augments the question prompts by retrieving and fusing Top-$K$ knowledge from the KVL, thereby enabling specialized Q\&A. When encountering new knowledge, RAG only needs to store the new knowledge in the repository to address new questions. Motivated by this, our proposed KA summarizes the data-specific knowledge from encountered samples into the knowledge vector library and then fuses them into the representation of input data in a weighted manner (Figure\textcolor{red}{~\ref{fig:motiv}}(b)). In such a way, there is no need to rely on expert knowledge to set the $K$ value as in RAG and the model can adaptively output a more fitting instance representation for different input data.

Furthermore, functional modules, such as KA in this paper should be optimized in each incremental session as demonstrated in previous works~\cite{Zhu_2021_CVPR,WANG2023111130,10168925}. However, the scarcity of training samples and the unavailable old data in incremental sessions make such training impossible. To address this issue, we propose a pseudo learning scheme of Incremental Pseudo Episode Learning (IPEL). The IPEL transfers the learned knowledge to incremental sessions by mimicking the real incremental setting. Concretely, because the data provided by the base session is sufficient, IPEL constructs a series of pseudo incremental tasks with the data sampled from the base session, which allows us to tune the knowledge to align with the context of subsequent incremental learning.

In summary, we make the following contributions in this paper:
\begin{itemize}
    \item We present an exploration of how to adapt the CLIP to the few-shot class incremental learning and contribute an effective framework to fine-tune the general knowledge of a CLIP to FSCIL. 
       
    \item We propose a knowledge adaptation network (KANet), where a knowledge adapter module is {designed} to tune the general knowledge to match the context of FSCIL, and a scheme of incremental pseudo episode learning (IPEL) is devised to transfer the learned knowledge to the incremental sessions. 

    \item We conduct comprehensive experiments on a broad range of datasets, including three previous FSCIL benchmark datasets and four new FSCIL benchmark datasets, and experimental results validate the efficacy of our proposed method.

\end{itemize}

\section{Related Work}\label{sec:related}

\subsection{Few-shot learning}
\label{subsec:fsl}
Few-shot learning (FSL) defines a task that aims to achieve fast adaptation to new classes using limited samples. To address this issue, recent works can be roughly divided into three groups, the metric-based, optimization-based, and hallucination-based methods. The metric-based methods ~\cite{Chikontwe_2022_CVPR,WU2022103349,MAI2019102781} focus on modeling discriminative relations between classifier weights and test features. For example, MAI \etal propose a method named Attentive Matching Network (AMN) that utilizes feature-level attention to learn discriminative inter-class relations. Unlike previous FSL methods, the optimization-based methods~\cite{pmlr-v70-finn17a,Jamal_2019_CVPR,Baik_2021_ICCV} focus on learning a satisfactory initialization for new class learning. For example, the classical optimization-based method MAML~\cite{pmlr-v70-finn17a} designs a two-steps optimization strategy which utilizes support samples in the inner loop and query samples in the outer loop to optimize the model. In such a way, MAML can help the model learn to use a few support samples to achieve fast adaptation for new classes in the inference stage. As the most intuitive and straightforward solution, the hallucination-based methods~\cite{Guo_2020_CVPR,Xu_2022_CVPR} generates fake samples or features to provide more training samples for new class learning. Despite differences existed in different methods, the common learning paradigm in FSL is to organize the training data in the form of meta task which is similar to the setting of inference task. As evidenced in existing FSL methods, such a learning paradigm can improve the model's generalization ability effectively. In this paper, our proposed Incremental Pseudo Episode Learning (IPEL) is motivated by this.

\subsection{Class-incremental learning}
\label{subsec:cil}

Class-incremental learning (CIL) aims to achieve incremental learning of new classes while retaining previously learned knowledge. The main challenge in CIL is the notorious catastrophic forgetting problem\cite{JODELET2022103582}. To address this issue, recent works can be divided into four groups, the rehearsal-based, regularization-based, isolation-based, and prompt-based methods. The regularization-based methods~\cite{lwf,PODNet} utilizes the knowledge distillation techniques~\cite{hinton2015distilling} to prevent the learned model from being over-optimized by new data. For example, PODNet~\cite{PODNet} proposes to distill features of each layer to prevent the catastrophic forgetting problem. Despite the effectiveness of regularization-based methods in maintaining the old knowledge, the model's plasticity is also constrained. Unlike these methods, the rehearsal-based methods~\cite{Rebuffi_2017_CVPR,wang2022memory} store and replay old exemplars to make the model retrospect old knowledge when learning new classes. The isolation-based methods~\cite{mallya2018piggyback,Yan_2021_CVPR} split the model into two parts, where one part is frozen to keep old knowledge and another part is trained to learn new knowledge. With the development of foundation models~\cite{radford2021learning,caron2021emerging}, the prompt-based methods~\cite{wang2022dualprompt,wang2022sprompts} adopt the foundation model as network pedestal to provide a satisfactory initialization for new class learning and utilizes prompt to learn new knowledge. The effectiveness of most CIL methods on new classes relies on tremendous data, but our proposed method only needs a few samples.

\subsection{Few-shot class-incremental learning}
\label{subsec:fscil}
FSCIL inherits the characteristics of CIL and FSL. As a research hotspot, Cheraghian\cite{Cheraghian_2021_CVPR} propose to distill the semantic information to mitigate the notorious catastrophic forgetting problem in FSCIL. Kang \etal \cite{kang2023on} propose a method dubbed SoftNet that splits the main model into two sub-networks based on their importance to old classes and only train the less important sub-network to learn new classes. Zhang\etal \cite{Zhang_2021_CVPR} propose a method that decouples the feature learning and classifier learning, where an adaption module is further devised to update the classifier based on the context between classifier weights. Similarly, Zhu\etal \cite{Zhu_2021_CVPR} propose a method that utilizes the relation between old classifier weights and new classifier weights to update the global classifier. Wang \etal \cite{10168925} argue that only updating classifier weights is insufficient and propose a method that updates both classifier weights and corresponding features. Zhang \etal \cite{Chi_2022_CVPR} devise a meta-learning scheme that mimics the multi-step incremental learning setting to construct pseudo incremental tasks and demonstrates such a scheme helps improve the model's plasticity. Hersche \etal \cite{hersche2022constrained} design an algorithm that stores old features and replays them to finetune a projection layer to output representative features. Most existing methods are built on the model trained from scratch. However, the efficacy of popular CLIP in solving FSCIL is rarely explored. In this paper, we focus on adapting the CLIP to FSCIL.

\subsection{Retrieval Augmented Generation}

In recent years, Large Language Models (LLMs)~\cite{openai2023gpt4,du2021glm} have shown stunning general content generation capabilities and have been applied to various downstream scenarios, such as chatbots. However, due to the lack of specialized domain knowledge, using naive LLMs can not effectively solve the relevant issues that exist in downstream tasks. To address this issue, various Retrieval Augmented Generation (RAG) methods are proposed to activate the power of Large Language Models (LLMs)~\cite{openai2023gpt4,du2021glm} to solve specialized tasks well. The core idea of RAG is to retrieve necessary knowledge from a specialized knowledge repository to enhance targets\cite{lewis2020retrieval}. For example, REALM~\cite{guu2020retrieval} and REPLUG\cite{shi2023replug} set the retriever to be trainable to effectively retrieve relevant knowledge. Asai \etal \cite{asai2023selfrag} argue that knowledge retrieved by previous RAG methods may not be helpful for response generation and propose a method that utilizes \textit{reflection} tokens to filter irrelevant knowledge. The designation of our Knowledge Adapter (KA) is inspired by these works.

\begin{figure*}
\centering
\includegraphics[width=1.7\columnwidth]{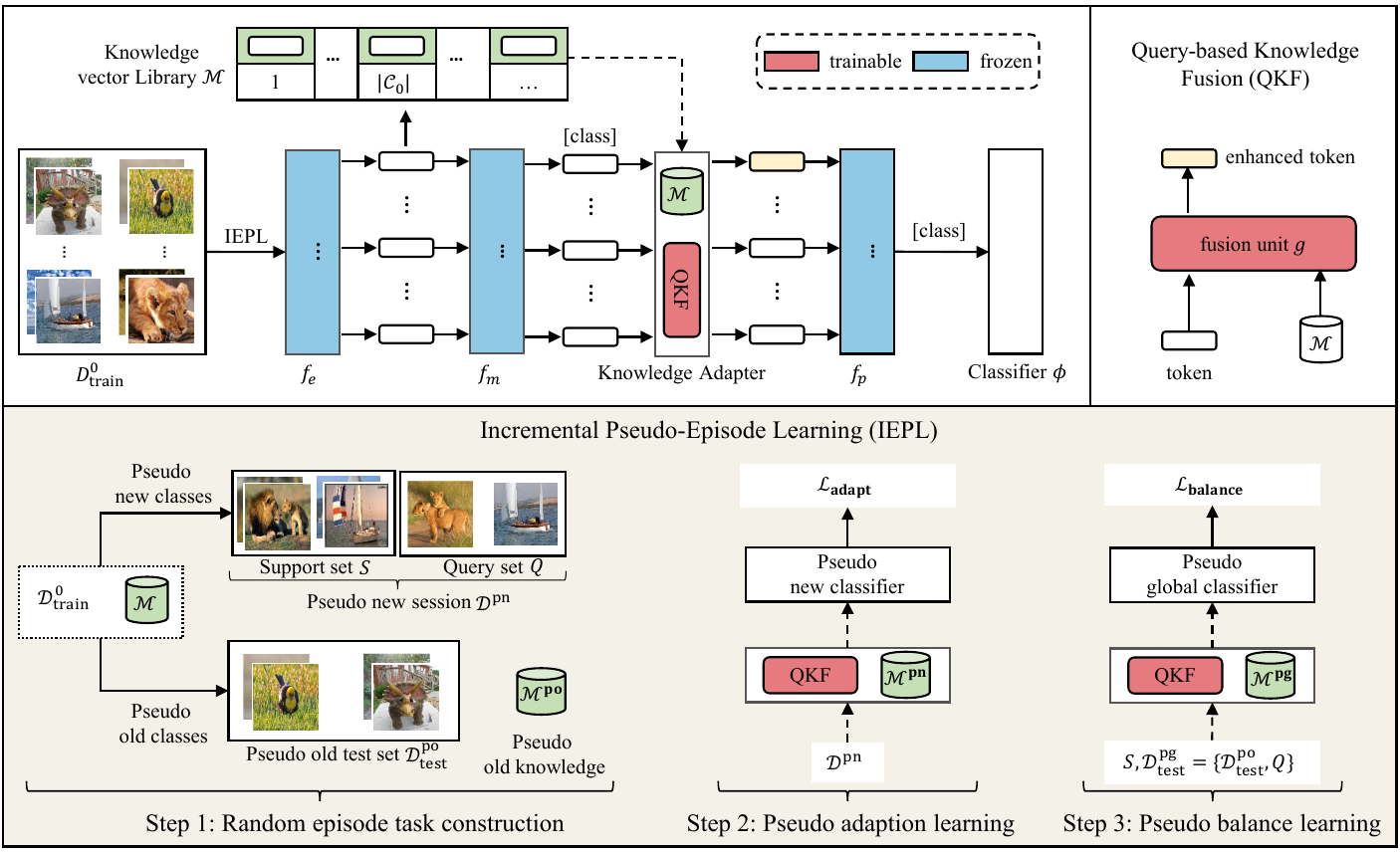}
\caption{Our proposed method adopts the pretrained image branch of CLIP as the backbone, where
 the knowledge adapter (KA) is plugged into one of the encoding layers and integrates data-specific knowledge stored in the knowledge vector library $\mathcal{M}$ and general knowledge of the CLIP using query-based knowledge fusion (QKF) to enhance instance representation. While the incremental pseudo episode learning (IPEL) scheme simulates real-world incremental settings and trains the KA via pseudo task adaption and balance learning, where $\mathcal{M}^{\text{pn}}$ refers to the pseudo new knowledge extracted from the support set $S$ and  $\mathcal{M}^{\text{pg}}$ indicates the pseudo global knowledge constructed by $\mathcal{M}^{\text{po}}$ and $\mathcal{M}^{\text{pn}}$.
\label{fig:method}}
\end{figure*}

\section{Preliminaries}\label{sec:problem}

Formally, let $\{\mathcal{D}^0, \mathcal{C}^{0}\}{\rightarrow}\{\mathcal{D}^1, \mathcal{C}^{1}\}{\rightarrow}...{\rightarrow}\{\mathcal{D}^i, \mathcal{C}^{i}\}(i>1)$ denote the data stream, where $\mathcal{D}^i$ and $\mathcal{C}^{i}$ denote the image data and label space of the $i$-th session. Each $\mathcal{D}^i$ consists of a training set $\mathcal{D}^{i}_{\text{train}}$ and a test set $\mathcal{D}^{i}_{\text{test}}$, where $\mathcal{D}^{i}_{\text{train}}$ and $\mathcal{D}^{i}_{\text{test}}$ satisfy $\mathcal{D}^{i}_{\text{train}}\cap{\mathcal{D}^{i}_{\text{test}}}=\emptyset$.
In different sessions, ${\mathcal{C}^{i}}\cap{\mathcal{C}^{j}}=\emptyset(i\neq{j})$. 
When the learning stage comes to the $i$-th session, only $\mathcal{D}^{i}_{\text{train}}$ {is available, but we need to evaluate our model using ${\mathcal{D}^{0}_{\text{test}}\cup\cdots\cup\mathcal{D}^{i}_{\text{test}}}(i\geq{0})$}. The specialty lies in FSCIL is that $\mathcal{D}^{0}_{\text{train}}$ {provides} sufficient training samples, while $\mathcal{D}^{j}_{\text{train}}(j>0)$ only includes a few training samples. For example, each class contained in $\mathcal{D}^{0}_{\text{train}}$ of the benchmark dataset CIFAR100 has 500 training sample, while that of $\mathcal{D}^{j}_{\text{train}}(j>0)$ has only 5 training samples. The imbalanced data distribution and the scarcity of training samples make the FSCIL challenging.

While being amazed by the excellent transfer ability of the CLIP, more and more researchers are devoting themselves to exploring the adaptation of the CLIP to downstream tasks, such as class-incremental learning~\cite{wang2022sprompts,wang2022dualprompt}, few-shot learning~\cite{zhang2022tip,zhang2023prompt}. However, though the CLIP can provide satisfactory initial representation for new classes which is extremely important in FSCIL, we surprisingly find such explorations for FSCIL are rare. In this paper, we fill the gap.
\section{Methodology}\label{sec:method}

To solve the challenging FSCIL, our proposed method employs the image branch\footnote{In this paper, we only use the image branch of CLIP. One major concern is that knowing the name of a new class may be unreality. In other words, we assume the CLIP has seen all defined classes in the world, the new emerging classes are all unknown. This assumption also coincides with the core idea of FSCIL. Additionally, without additional remarks, we use CLIP to denote the image branch of CLIP for simplicity.} of CLIP as the network pedestal and fuses the task-specific knowledge into the CLIP to make the CLIP output more fitting features.  Figure~\ref{fig:method} depicts the proposed framework, we deploy a Knowledge Adapter (KA, Section~\ref{sec:KA}) module to learn the task-specific knowledge, where the KA consists of a knowledge vector library and a query-based knowledge fusion mechanism. In KA, the knowledge vector library summarizes the data-specific knowledge from available data. Next, the query-based knowledge fusion integrates the resulting knowledge into the middle feature to achieve the refinement of the encoder's output. Furthermore, since the number of training samples is limited and the old data is not available in incremental sessions, the KA's parameters are optimized using the Incremental Pseudo Episode Learning (IPEL, Section~\ref{sec:IPEL}) to foster the transfer of learned knowledge to the incremental sessions. During the incremental stage, the following steps are taken:
\emph{(1)} summarize knowledge from the training data into the knowledge vector library,
\emph{(2)} use the query-based knowledge fusion mechanism to refine the representations of inputs,
\emph{(3)} obtain class-wise mean features of the refined training features and concatenate them with the old classifier weights to construct a new classifier, 
and \emph{(4)} for each test sample, refine the instance's representation using KA, and then apply the new classifier to make predictions.

\subsection{{Knowledge Adapter}}
\label{sec:KA}
Let $f={f_e}\circ{f_m}\circ{f_p}\circ{\phi}$ denotes the image branch of CLIP, where $f_e$, $f_m$, and $f_p$ respectively represent early-stage, middle-stage, and post-stage encoding layers \footnote{In default, we treat the former 4 layers of ViT as the early-stage encoding layers,  the subsequent 6 layers are taken as the middle-stage encoding layers, and the remaining layers except the classification head as the post-stage encoding layers. The discussion regarding this split setting can be seen in Section \ref{subsubsec:layer_conf}.}, $\phi$ indicates the classifier parameterized by ${\theta}_c$.

\noindent\textbf{Knowledge vector library} is used to summarize the data-specific knowledge from the training data, which is a preliminary step for the following query-based knowledge fusion. Concretely, we first input a training image $x\in{\mathbb{R}^{C\times{H}\times{W}}}$ to the $f_e$ and get the embedding feature $x_e=f_e(x)\in{\mathbb{R}}^{(L+1)\times{D}}$ constituted by a series of tokens with the length $L+1$, where the first token $x_e^0\in{\mathbb{R}}^{1\times{D}}$ is often called the \texttt{[class]} token, the rests are called the \texttt{[patch]} tokens. Then, because the \texttt{[class]} token captures the instance's global information, a straightforward method to construct the knowledge vector library is to use $x_e^0$. However, using the instance-level \texttt{[class]} token will consume much memory. To reduce the memory consumption, we compute the class-level \texttt{[class]} token by averaging the instance-level \texttt{[class]} tokens that belong to the same class. Finally, the knowledge vector library $\mathcal{M}$ is constructed as follows:
\begin{equation}
    \mathcal{M} = [\mathcal{M}^1, ..., \mathcal{M}^{|\mathcal{C}|}]\in{\mathbb{R}}^{|\mathcal{C}|\times{D}},
\end{equation}
where $|\mathcal{C}|$ refers to the number of seen class, $\mathcal{M}^{|\mathcal{C}|}$ represents the class-level \texttt{[class]} token of class $|\mathcal{C}|$.

During incremental sessions, we extract the class knowledge from the available training data following the above steps, and the codebook can be analogously constructed for future knowledge fusion.

\noindent\textbf{{Query-based knowledge fusion}} fuses the data-specific knowledge contained in the knowledge vector library $\mathcal{M}$ into the instance's middle feature to achieve the refinement of the model's output. Concretely, given the embedding feature $x_e$. We first input $x_e$ into $f_m$ and get the embedding feature $x_m=f_m(x_e)=\{x_m^0, ..., x_m^L\}\in{\mathbb{R}}^{(L+1)\times{D}}$. Similarly, $x_m^0\in{\mathbb{R}}^{1\times{D}}$ refers to the encoded \texttt{[class]} token. Then, we use the knowledge of $\mathcal{M}$ to enhance $x_m^0$ as follow:
\begin{equation}
\label{eq:fuse}
    \hat{x}_m^0 = g({x_m^0}, \mathcal{M};{\theta}_g)\in{\mathbb{R}}^{1\times{D}},
\end{equation}
where $\hat{x}_m^0$ is the enhanced $x_m^0$, $g$ refers to the fusion unit, and ${\theta}_g$ indicates the parameters of $g$ used to learn the task-specific knowledge. Given the powerful information interaction capability of the Transformer~\cite{NIPS2017_3f5ee243}, in this paper, we adopt the Transformer as the fusion unit $g$, and set $x_m^0$ as the \texttt{query} input, $ \mathcal{M}$ as the \texttt{key} and \texttt{value} inputs as shown in Figure~\ref{fig:method}. With $\hat{x}_m^0$, we next input the enhanced embedding feature $\hat{x}_m=\{\hat{x}_m^0, ..., x_m^L\}\in{\mathbb{R}}^{(L+1)\times{D}}$ to $f_p$ and get the refined feature $x_p=f_p(\hat{x}_m)=\{x_p^0, ..., x_p^L\}\in{\mathbb{R}}^{(L+1)\times{D}}$. Finally, we select the $x_p^0$ as the final refined global representation of $x$. To facilitate the following descriptions, we replace the notation $x_p^0$ with $f(x,\mathcal{M})$, which means the representation of $x$ is refined by $\mathcal{M}$.

\noindent\textbf{Advantages of Knowledge Adapter}{ lies in the effective fusion of multiple types of knowledge for instance representation refinement. Particularly, the frozen weights of the CLIP take along the knowledge from large-scale data, while the data-specific knowledge is summarized into the knowledge vector library $\mathcal{M}$, and the task-specific knowledge will be learned in the introduced projection matrices. With the knowledge query and fusion procedures (Eq.~\ref{eq:fuse}), the above types of knowledge can be fused into instance features, such that a more comprehensive representation can be harvested to benefit the following learning tasks.}    

\subsection{{Incremental Pseudo Episode learning}}
\label{sec:IPEL}

Unlike typical class incremental learning giving ample data in incremental stages, FSCIL struggles to support adjustments to previously learned knowledge with scarce samples after running into the incremental stage. To surmount this challenge, incremental pseudo episode learning (IPEL) simulates incremental learning using the base session's data, aiming to smooth the acquired knowledge for the incremental stage and prime the actual incremental learning. In IPEL, there are three main steps: random episode task construction, pseudo adaptation learning, and pseudo balance learning.

\noindent\textbf{Random episode task construction}
\label{tc}
builds a series of pseudo learning tasks using the data in the base session for knowledge transfer. Similar to the actual incremental learning, pseudo tasks consist of new classes' data, including a support set and query set, as well as the test set of old classes. These data components are constructed as 
 follows:  
\emph{(i) Pseudo new session }$\mathcal{D}^{\text{pn}}$.
Formally, we sample two disjoint datasets from $\mathcal{D}^0_{\text{train}}$ to construct the support set $S$ and query set $Q$ with the well-known $N$-way-$K$-shot setting, which means $N$ classes are randomly sampled from $\mathcal{C}^0$ as the pseudo new classes and $K$ samples for each class are randomly selected from $\mathcal{D}^0_{\text{train}}$ to construct the support set $S$. In IPEL, $S$ serves as the pseudo training set while $Q$ serves as the pseudo test set.
\emph{(ii) Pseudo test set of old classes }$\mathcal{D}^{\text{po}}_{\text{test}}$.
To balance plasticity and stability, we consider the remaining classes in $\mathcal{C}^0$ as the pseudo old classes. 
Then, among pseudo old classes, we randomly select several samples from $\mathcal{D}^0_{\text{train}}$ as the test set $\mathcal{D}^{\text{po}}_{\text{test}}$ of pseudo old classes.
\emph{(iii) Pseudo old knowledge }$\mathcal{M}^{\text{po}}$. Given the known pseudo old classes, we select their corresponding knowledge from  $\mathcal{M}$ to construct $\mathcal{M}^{\text{po}}$.

Overall, for each episode, we use the combination $ \{\mathcal{D}^{\text{pn}}, \mathcal{D}^{\text{po}}_{\text{test}}, \mathcal{M}^{\text{po}}\}$ to form a pseudo learning task.

\noindent\textbf{Pseudo adaptation learning}
\label{pal}
acts on the constructed pseudo tasks to propagate and smooth the historical knowledge to match the context of few-shot class-incremental learning, enabling the subsequent few-shot incremental tasks to be warmed up and executed with greater ease. Concretely,  we first summarise the pseudo new knowledge $\mathcal{M}^{\text{pn}}$ from $S$ with the step described in Section~\ref{sec:KA}. Then, we input the $S$ and $Q$ to the encoder and utilize the query-based knowledge fusion to get the refined features $f(x^s,\mathcal{M}^{\text{pn}})$ and $f(x^q,\mathcal{M}^{\text{pn}})$ of $S$ and $Q$, where $x^s \in S$ and $x^q \in Q$. Next, we compute the prototypes of pseudo new classes by averaging $f(x^s,\mathcal{M}^{\text{pn}})$ that belong to the same class and use the computed prototypes to initialize pseudo new classifier weights ${\theta}^{\text{pn}}\in{\mathbb{R}}^{N\times{D}}$. After that, we employ ${\theta}^{pn}$ to make predictions for ${f}(x^q,\mathcal{M}^{\text{pn}})$ as follows:
\begin{equation}
\label{eq:prob}
    P^q = \text{softmax} (\alpha<{f}(x^q,\mathcal{M}^{\text{pn}}), {\theta}^{\text{pn}}>),
\end{equation}
where $P^{q}$ denotes the prediction results, $\alpha$ refers to the scaling factor, $<a,b>=\frac{a\cdot{b}}{||a||_2{||b||_2}}$ indicates the cosine similarity. In the end, we compute the adaptation loss $\mathcal{L}_{\text{adapt}}$ by
\begin{equation}
\label{loss_1}
    \mathcal{L}_{\text{adapt}} = \mathcal{L}_{\text{ce}}(P^{q}, Y^{q}),
\end{equation}
where $Y^{q}$ denotes the labels of $Q$.

\noindent\textbf{Pseudo balance learning} addresses the gap in pseudo adaptation learning approaches as it mostly prioritizes the pseudo new classes and pays rare attention to the remaining pseudo old classes.
\label{pbl} 
Particularly, we first combine $\mathcal{D}^{\text{po}}_{\text{test}}$ and $Q$ to constitute a pseudo global test set $\mathcal{D}^{\text{pg}}_{\text{test}}$ and concatenate the $\mathcal{M}^{\text{po}}$ and $\mathcal{M}^{\text{pn}}$ to construct the pseudo global knowledge $\mathcal{M}^{\text{pg}}$. 
Next, we feed $\mathcal{D}^{\text{pg}}_{\text{test}}$ and $S$ into the encoder to obtain their refined representations $f(x^{\text{pg}}, \mathcal{M}^{\text{pg}})$, $f(x^s, \mathcal{M}^{\text{pg}})$, where $x^{\text{pg}} \in \mathcal{D}^{\text{pg}}_{\text{test}}$.
Then, we compute the prototypes of $S$ by averaging $f(x^s, \mathcal{M}^{\text{pg}})$ that belong to the same class. 
After that, we concatenate the computed prototypes and the pseudo old classifier weights ${\theta}^{\text{po}}$ selected from ${\theta}_c$ to initialize a pseudo global classifier ${\theta}^{{\text{pg}}}$. Next, we use ${\theta}^{{\text{pg}}}$ and Eq.\ref{eq:prob} to compute the prediction results $P^{\text{pg}}$ of $f(x^{\text{pg}}, \mathcal{M}^{\text{pg}})$. In the end, we compute the balance loss $\mathcal{L}_{\text{balance}}$ as below: 
\begin{equation}
\label{loss_1}
    \mathcal{L}_{\text{balance}} = \mathcal{L}_{ce}(P^{\text{pg}}, Y^{\text{pg}}),
\end{equation}
where $Y^{\text{pg}}$ denotes the corresponding labels of $\mathcal{D}^{\text{pg}}$.

Overall, the full objective is given by:
\begin{equation}
\label{loss_total}
    \mathcal{L} = {\lambda}_{\text{adapt}}\mathcal{L}_{\text{adapt}} + {\lambda}_{\text{balance}}\mathcal{L}_{\text{balance}},
\end{equation}
where ${\lambda}_{\text{adapt}}$ and ${\lambda}_{\text{balance}}$ are hyper-parameters to trade-off two losses.
\section{Experiments}\label{sec:exp}

\subsection{Datasets}\label{subsec:data}
\textbf{CIFAR100.} The CIFAR100 \cite{cifar} dataset consists of 100 classes, {where each dataset consists of 500 training samples and 100 test samples. We follow \cite{Tao_2020_CVPR} to split into 1 base session and 8 incremental sessions, where the base session includes 60 classes and each incremental session has 5 classes. Further, except for the base session, we only sample 5 training samples for each class in incremental sessions.}

\textbf{Caltech-UCSD Birds-200-2011.} The CUB200~\cite{cub} dataset consists of 11,788 images from 200 classes. {We follow \cite{Tao_2020_CVPR} to construct 11 learning sessions, \ie 1 base session and 10 incremental sessions, where the base session includes 100 classes and each incremental session consists of 10 classes. Similarly, except for the base session, each class in incremental sessions only be provided with 5 training samples.}

\textbf{ImageNet-R.}
Previous FSCIL methods often adopt \textit{mini}ImageNet, the subset of ImageNet~\cite{2015ImageNet}, as the third benchmark dataset. However, this dataset has a large overlap with the large-scale data used in the CLIP. To make a convincing evaluation, we follow the practice in class-incremental learning~\cite{wang2022learning} and adopt the ImageNet-R as the third benchmark. This dataset contains various renditions of 200 ImageNet classes resulting in 30,000 RGB images~\cite{hendrycks2021many}. {We adopt the same incremental setting as in CUB200 to construct 11 learning sessions}.

\begin{table*}[ht]\footnotesize
\begin{center}
\caption{Comparison with previous methods on CIFAR100, where $^{\dagger}$ denotes our reproduced result, $*$ indicates results copied from~\cite{Tao_2020_CVPR}. Methods (top-down) are sorted in ascending order based on the average accuracy, and the best results under different backbones are highlighted in bold. Our proposed method achieves the best performance on each session.} 
\label{tab:cmp_sota_cifar}
    \setlength{\tabcolsep}{2.7mm}{
\begin{tabular}{l|c|ccccccccc|cc}
\toprule
\multirow{2}{*}{Method} & \multirow{2}{*}{Enc.} & \multicolumn{9}{c|}{sessions(1+8)} & \multirow{2}{*}{\texttt{Avg.}} & \multirow{2}{*}{\texttt{PD}}\\
\cmidrule{3-11}
{} & {} & 0 & 1 & 2 & 3 & 4 & 5 & 6 & 7 & 8 & {} & {}  \\
\midrule
\midrule
NCM$^{*}$ \cite{Hou_2019_CVPR}         & {\multirow{15}{*}{RN}} & 64.10 & 53.05 & 43.96 & 36.97 & 31.61 & 26.73 & 21.23 & 16.78 & 13.54 & 34.22 & 50.56 \\
iCaRL$^{*}$ \cite{Rebuffi_2017_CVPR}   & {} & 64.10 & 53.28 & 41.69 & 34.13 & 27.93 & 25.06 & 20.41 & 15.48 & 13.73 & 32.87 & 50.37 \\
EEIL$^{*}$ \cite{Castro_2018_ECCV}     & {} & 64.10 & 53.11 & 43.71 & 35.15 & 28.96 & 24.98 & 21.01 & 17.26 & 15.85 & 33.79 & 48.25 \\
TOPIC\cite{Tao_2020_CVPR}              & {} & 64.10 & 55.88 & 47.07 & 45.16 & 40.11 & 36.38 & 33.96 & 31.55 & 29.37 & 42.62 & 34.73 \\
SPPR\cite{Zhu_2021_CVPR}               & {} & 63.97 & 65.86 & 61.31 & 57.60 & 53.39 & 50.93 & 48.27 & 45.36 & 43.32 & 54.45 & 20.65 \\
F2M\cite{shi2021overcoming}            & {} & 71.45 & 68.10 & 64.43 & 60.80 & 57.76 & 55.26 & 53.53 & 51.57 & 49.35 & 59.14 & 22.10 \\
CEC\cite{Zhang_2021_CVPR}              & {} & 73.07 & 68.88 & 65.26 & 61.19 & 58.09 & 55.57 & 53.22 & 51.34 & 49.14 & 59.53 & 23.93 \\
MCNet~\cite{MCNet}                     & {} & 73.30 & 69.34 & 65.72 & 61.70 & 58.75 & 56.44 & 54.59 & 53.01 & 50.72 & 60.40	& 22.58 \\
CLOM~\cite{NEURIPS2022_ae817e85}       & {} & 74.20 & 69.83 & 66.17 & 62.39 & 59.26 & 56.48 & 54.36 & 52.16 & 50.25	& 60.57 & 23.95 \\
MetaFSCIL\cite{Chi_2022_CVPR}          & {} & 74.50 & 70.10 & 66.84 & 62.77 & 59.48 & 56.52 & 54.36 & 52.56 & 49.97 & 60.79 & 24.53 \\
MFS3~\cite{XU2023110394}               & {} & 73.42	& 69.85	& 66.44	& 62.81	& 59.78	& 56.94	& 55.04	& 53.00	& 51.07	& 60.93 & 22.35 \\
C-FSCIL\cite{hersche2022constrained}   & {} & 77.47 & 72.40 & 67.47 & 63.25 & 59.84 & 56.95 & 54.42 & 52.47 & 50.47 & 61.64	& 27.00 \\
FACT\cite{Zhou_2022_CVPR}              & {} & 74.60 & 72.09 & 67.56 & 63.52 & 61.38 & 58.36 & 56.28 & 54.24 & 52.10 & 62.24 & 22.50 \\
TEEN~\cite{wang2023fewshot}            & {} & 74.92 & 72.65 & 68.74 & 65.01 & 62.01 & 59.29 & 57.90 & 54.76 & 52.64 & 63.10 & 22.28 \\ 
ALICE~\cite{ALICE}                     & {} & 79.00 & 70.50 & 67.10 & 63.40 & 61.20 & 59.20 & 58.10 & 56.30 & 54.10 & 63.21 & 24.90 \\
SoftNet~\cite{kang2023on}              & {} & 79.88 & 75.54 & 71.64 & 67.47 & 64.45 & 61.09 & 59.07 & 57.29 & 55.33 & 65.75 & 24.55 \\
WaRP~\cite{WaRP}                       & {} & 80.31 & 75.86 & 71.87 & 67.58 & 64.39 & 61.34 & 59.15 & 57.10 & 54.74 & 65.82 & 25.57 \\
CaBD~\cite{CaBD}                       & {} & 79.45 & 75.38 & 71.84 & 67.95 & 64.96 & 61.95 & 60.16 & 57.67 & 55.88 & 66.14 & 23.57  \\
NC-FSCIL~\cite{yang2023neural}         & {} & 82.52 & 76.82 & 73.34 & 69.68 & 66.19 & 62.85 & 60.96 & 59.02 & 56.11 & 67.50 & 26.41 \\
CEC+~\cite{10168925}                   & {} & {81.25} & 77.23 & 73.30 & 69.41 & 66.69 & 63.93 & 62.16 & 59.62 & 57.41 & {67.89} & 23.84 \\
\textbf{KANet(Ours)}                   & {} & 79.53 & 76.20 & 73.01 & 69.15 & 67.23 & 64.96 & 63.97 & 62.01 & 59.62 & \textbf{68.41} & \textbf{19.91} \\

\midrule
Finetune$^{\dagger}$                   & \multirow{4}{*}{ViT}  & 85.67 & 81.14 & 75.37 & 59.68 & 50.31 & 24.00 & 21.03 & 16.29 & 16.85 & 47.82 & 68.82 \\
Joint$^{\dagger}$                      & {} & 85.67 & 79.77 & 75.06 & 70.84 & 67.11 & 64.06 & 61.88 & 59.49 & 56.67 & 68.95 & 29.00 \\
CEC+$^{\dagger}$~\cite{10168925}       & {} & 85.67 & 78.55 & 76.51 & 73.80 & 72.92 & 71.67 & 71.76 & 70.55 & 68.90 & 74.48 & 16.77 \\
\textbf{KANet(Ours)}                   & {} & 85.67 & 79.94 & 78.06 & 75.43 & 74.43 & 73.11 & 73.16 & 71.95 & 70.22 & \textbf{75.77} & \textbf{15.45} \\
\bottomrule 

\end{tabular}}

\end{center}
\end{table*}

\begin{table*}[ht]\footnotesize
\begin{center}
\caption{Comparison with previous methods on CUB200, where $^{\dagger}$ denotes our reproduced result, $*$ indicates results copied from~\cite{Tao_2020_CVPR}. Methods (top-down) are sorted in ascending order based on the average accuracy, and the best results under different backbones are highlighted in bold. Our proposed method achieves the best performance on almost each session.} 
\label{tab:cmp_sota_cub_imr}
    \setlength{\tabcolsep}{1.85mm}{
\begin{tabular}{l|c|ccccccccccc|cc}
\toprule
\multirow{2}{*}{Method} & \multirow{2}{*}{Enc.}  & \multicolumn{11}{c|}{sessions(1+10)} & \multirow{2}{*}{\texttt{Avg.}} & \multirow{2}{*}{\texttt{PD}}\\
\cmidrule{3-13}
{} & {} & 0 & 1 & 2 & 3 & 4 & 5 & 6 & 7 & 8  & 9  & 10 & {} & {}  \\
\midrule
\midrule
NCM$^{*}$ \cite{Hou_2019_CVPR}       &  \multirow{18}{*}{RN}  & 68.68 & 57.12 & 44.21 & 28.78 & 26.71 & 25.66 & 24.62 & 21.52 & 20.12 & 20.06 & 19.87 & 32.49 & 48.81 \\
EEIL$^{*}$~\cite{Castro_2018_ECCV}   & {} & 68.68 & 53.63 & 47.91 & 44.20 & 36.30 & 27.46 & 25.93 & 24.70 & 23.95 & 24.13 & 22.11 & 36.27 & 46.57 \\
iCaRL$^{*}$~\cite{Rebuffi_2017_CVPR} & {} & 68.68 & 52.65 & 48.61 & 44.16 & 36.62 & 29.52 & 27.83 & 26.26 & 24.01 & 23.89 & 21.16 & 36.67 & 47.52 \\
TOPIC~\cite{Tao_2020_CVPR}           & {} & 68.68 & 62.49 & 54.81 & 49.99 & 45.25 & 41.40 & 38.35 & 35.36 & 32.22 & 28.31 & 26.28 & 43.92 & 42.40 \\
SPPR~\cite{Zhu_2021_CVPR}            & {} & 68.68 & 61.85 & 57.43 & 52.68 & 50.19 & 46.88 & 44.65 & 43.07 & 40.17 & 39.63 & 37.33 & 49.34 & 31.35 \\
CEC~\cite{Zhang_2021_CVPR}           & {} & 75.85 & 71.94 & 68.50 & 63.50 & 62.43 & 58.27 & 57.73 & 55.81 & 54.83 & 53.52 & 52.28 & 61.33 & 23.57 \\
MetaFSCIL~\cite{Chi_2022_CVPR}       & {} & 75.90 & 72.41 & 68.78 & 64.78 & 62.96 & 59.99 & 58.30 & 56.85 & 54.78 & 53.82 & 52.64 & 61.93 & 23.26 \\
MFS3~\cite{XU2023110394}             & {} & 75.63 & 72.51	& 69.65	& 65.29	& 63.13	& 60.38	& 58.99	& 57.41	& 55.55	& 54.95	& 53.47	& {62.45} & {22.16} \\
F2M~\cite{shi2021overcoming}         & {} & 77.13 & 73.92 & 70.27 & 66.37 & 64.34 & 61.69 & 60.52 & 59.38 & 57.15 & 56.94 & 55.89 & 63.96 & 21.24 \\
FACT~\cite{Zhou_2022_CVPR}           & {} & 75.90 & 73.23 & 70.84 & 66.13 & 65.56 & 62.15 & 61.74 & 59.83 & 58.41 & 57.89 & 56.94 & 64.42 & 18.96 \\
WaRP~\cite{WaRP}                     & {} & 77.74 & 74.15 & 70.82 & 66.90 & 65.01 & 62.64 & 61.40 & 59.86 & 57.95 & 57.77 & 57.01 & 64.66 & 20.73 \\
SoftNet~\cite{kang2023on}            & {} & 78.07 & 74.58 & 71.37 & 67.54 & 65.37 & 62.60 & 61.07 & 59.37 & 57.53 & 57.21 & 56.75 & 64.68 & 21.32 \\
MCNet~\cite{MCNet}                   & {} & 77.57 & 73.96 & 70.47 & 65.81 & 66.16 & 63.81 & 62.09 & 61.82 & 60.41 & 60.09 & 59.08 & 65.57 & 18.49 \\
ALICE~\cite{ALICE}                   & {} & 77.40 & 72.70 & 70.60 & 67.20 & 65.90 & 63.40 & 62.90 & 61.90 & 60.50 & 60.60 & 60.10 & 65.75 & \textbf{17.30} \\
TEEN~\cite{wang2023fewshot}          & {} & 77.26 & 76.13 & 72.81 & 68.16 & 67.77 & 64.40 & 63.25 & 62.29 & 61.19 & 60.32 & 59.31 & 66.27 & 18.13 \\
CLOM~\cite{NEURIPS2022_ae817e85}     & {} & 79.57 & 76.07 & 72.94 & 69.82 & 67.80 & 65.56 & 63.94 & 62.59 & 60.62 & 60.34 & 59.58 & 67.17 & 19.99 \\
NC-FSCIL~\cite{yang2023neural}       & {} & 80.45 & 75.98 & 72.30 & 70.28 & 68.17 & 65.16 & 64.43 & 63.25 & 60.66 & 60.01 & 59.44 & 67.28 & 21.01 \\
CaBD~\cite{CaBD}                     & {} & 79.12 & 75.37 & 72.80 & 69.05 & 67.53 & 65.12 & 64.00 & 63.51 & 61.87 & 61.47 & 60.93 & 67.34 & 18.19 \\
CEC+~\cite{10168925}                 & {} & 79.46 & 76.11 & 73.12 & 69.31 & 67.97 & 65.86 & 64.50 & 63.83 & 62.20 & 62.00 & 60.97 & 67.76 & {18.49} \\
\textbf{KANet(Ours)}                          & {} & 81.16 & 78.22 & 75.28 & 71.49 & 71.02 & 68.73 & 67.24 & 65.99 & 64.29 & 63.96 & 63.49 & \textbf{70.07} & 17.67 \\

\midrule
Finetune$^{\dagger}$   & \multirow{4}{*}{ViT} & 82.00 & 76.72 & 70.42 & 60.70 & 45.24 & 25.75 & 21.39 & 16.84 & 13.05 & 11.34 & 10.39 & 39.44 & 71.61 \\
Joint$^{\dagger}$                   & {} & 82.00 & 78.48 & 75.10 & 72.54 & 70.31 & 68.27 & 66.71 & 65.95 & 63.85 & 64.66 & 63.19 & 70.10 & 18.81 \\
CEC+$^{\dagger}$~\cite{10168925}    & {} & 82.00 & 76.68 & 74.97 & 72.27 & 71.37 & 69.89 & 68.94 & 68.38 & 66.89 & 67.48 & 67.12 & 71.45 & 15.88 \\
\textbf{KANet(Ours)}                         & {} & 82.00 & 77.99 & 76.68 & 74.25 & 73.37 & 71.55 & 70.66 & 70.26 & 69.13 & 69.65 & 69.35 & \textbf{73.17} & \textbf{12.65} \\

\bottomrule 

\end{tabular}}

\end{center}
\end{table*}

\begin{table*}[ht]\footnotesize
\begin{center}
\caption{Comparison with previous methods on ImageNet-R, where $^{\dagger}$ denotes our reproduced result, $*$ indicates results copied from~\cite{Tao_2020_CVPR}. Methods (top-down) are sorted in ascending order based on the average accuracy, and the best results are highlighted in bold. Our proposed method achieves the best performance on almost all sessions.} 
\label{tab:cmp_sota_cub_imr}
    \setlength{\tabcolsep}{2.15mm}{
\begin{tabular}{l|ccccccccccc|cc}
\toprule
\multirow{2}{*}{Method} & \multicolumn{11}{c|}{sessions (1+10)} & \multirow{2}{*}{\texttt{Avg.}} & \multirow{2}{*}{\texttt{PD}}\\
\cmidrule{2-12}
{} & 0 & 1 & 2 & 3 & 4 & 5 & 6 & 7 & 8  & 9  & 10 & {} & {}  \\
\midrule
Finetune$^{\dagger}$               & 78.27 & 73.06 & 66.94 & 55.46 & 38.98 & 28.67 & 20.21 & 16.16 & 13.7 & 10.77 & 8.95 & 37.38 & 69.32 \\ 
Joint$^{\dagger}$                  & 78.27 & 74.03 & 69.78 & 66.49 & 62.9 & 61.18 & 58.75 & 57.10 & 55.76 & 54.30 & 52.65 & {62.84} & {25.62} \\
CEC+$^{\dagger}$~\cite{10168925}   & 78.27 & 73.52 & 71.94 & 70.26 & 68.21 & 66.42 & 65.19 & 64.14 & 63.91 & 62.58 & 61.60 & 67.82 & 16.67 \\
\textbf{KANet(Ours)}               & 78.27 & 74.58 & 73.11 & 71.51 & 69.48 & 67.67 & 66.21 & 65.27 & 65.02 & 63.91 & 62.87 & {\textbf{68.91}} & {\textbf{15.40}} \\
\bottomrule 

\end{tabular}}

\end{center}
\end{table*}

\subsection{Implementation Details}
\label{subsec:impl}
\noindent\textbf{Architecture and Training.} 
{We adopt PyTorch~\cite{paszke2019pytorch} to implement our proposed method and} the image branch of CLIP-ViT-B/32\cite{radford2021learning} as our backbone. Our knowledge adapter is implemented using a Transformer block and plugged into the 10th encoder layer, where the knowledge stored in the knowledge vector library is summarized from the 4th encoder layer. We set the maximum epoch to 50 and adopt Adam as the optimizer, where the learning rate starts from 0.03 and decays with cosine annealing. Following~\cite{Zhu_2022_CVPR}, we resize all images to 224$\times$224. Random resized crop, random horizontal flip, and color jitter are employed to preprocess the data. The scaling factor $\alpha$, the ${\lambda}_{adapt}$ and ${\lambda}_{balance}$ are set to 16.0, 1.5, and 2.0, respectively. At the incremental pseudo episode learning stage, we randomly sample 200 pseudo incremental learning tasks for each epoch, where each pseudo incremental learning task is constructed with the 20-way-10-shot sampling setting, which means we randomly sample 20 classes from {the label space of base session} as pseudo new classes and 10 images for each {sampled} pseudo new class to construct the support set $S$. As for the query set $Q$, 15 images for each pseudo new class are randomly sampled. To construct the pseudo test set of old classes $\mathcal{D}^{po}_{test}$, we directly sample 128 images from the training data of pseudo old classes for computation efficiency. Though we mainly focus on how to adopt the CLIP to the context of FSCIL, we also adapt our proposed method to ResNet-18~\cite{ResNet} for a fair comparison, where we use the data-specific knowledge provided by CLIP and plug the knowledge adapter in the last layer. More concretely, the data-specific knowledge is summarized from the 4th encoder layer of the CLIP, then a projection layer with a hidden dim of 2048 to map the feature dimension (768) of CLIP to that (512) of ResNet-18.

\noindent\textbf{Evaluation Metrics.}
We use the average accuracy \texttt{Avg.}$=\frac{1}{n+1}\sum_{i=0}^{n}\mathcal{A}_{i}$ to measure overall performance across sessions, and the performance drop rate \texttt{PD}$= \mathcal{A}_0-\mathcal{A}_{n}$ to quantify forgetting as~\cite{Zhang_2021_CVPR}, where $\mathcal{A}_0$ and $\mathcal{A}_n$ refers to the accuracy of the first session and the last session.

\subsection{Quantitative Comparisons}
\label{subsec:cmp}
To validate the effectiveness of our proposed method, we conduct comprehensive comparisons with previous methods, including, {previous} class-incremental learning methods (iCaRL~\cite{Rebuffi_2017_CVPR}, EEIL~\cite{Castro_2018_ECCV}, and NCM~\cite{Hou_2019_CVPR}) and few-shot class-incremental learning (FSCIL) methods (TOPIC \cite{Tao_2020_CVPR}, SPPR~\cite{Zhu_2021_CVPR}, F2M~\cite{shi2021overcoming}, CEC~\cite{Zhang_2021_CVPR}, CLOM~\cite{NEURIPS2022_ae817e85}, C-FSCIL~\cite{hersche2022constrained}, MetaFSCIL~\cite{Chi_2022_CVPR}, FACT~\cite{Zhou_2022_CVPR}, ALICE~\cite{ALICE}, MCNet~\cite{MCNet}, SoftNet~\cite{kang2023on}, TEEN~\cite{wang2023fewshot}, NC-FSCIL~\cite{yang2023neural}, WaRP~\cite{WaRP}, CaBD~\cite{CaBD} and CEC+~\cite{10168925}). For a fair comparison, we reproduce CEC+, the state-of-the-art method, using the same ViT/B-32 backbone of CLIP.
As shown in Table \ref{tab:cmp_sota_cifar} and \ref{tab:cmp_sota_cub_imr}, we can see that:
\begin{itemize}
    \item On CIFAR100, our proposed method not only achieves the highest \texttt{Avg.}, but also the smallest \texttt{PD} value. Particularly, using ViT as the backbone, our proposed method outperforms the second-best method CEC+ by a margin of 1.29$\%$ on \texttt{Avg.} and 1.32 $\%$ on the \texttt{PD.} value. 
    \item On CUB200, our proposed method {achieves the highest \texttt{Avg.} and smallest \texttt{PD.}}. Particularly, {when using \texttt{Avg.} to evaluate the performance, our method surpasses CEC+ 1.72$\%$. When using \texttt{PD.} as a metric, our method achieves an improvement of 3.23 $\%$ over CEC+.}
    \item On CIFAR100 and CUB200, using ResNet as the backbone, our proposed method still achieves consistent improvement over CEC+ and other methods.
    \item On ImageNet-R, compared to CEC+, our proposed method achieves an improvement of 1.09$\%$ on \texttt{Avg.} and 1.27 $\%$ on the \texttt{PD.}.
    \item Additionally, the largest \texttt{PD.} value given by the Finetune method on each dataset indicates that training with limited samples makes the model suffer from the notorious catastrophic forgetting and overfitting problems. Even when using the joint training method, an upper bound in class-incremental learning, still can not yield satisfactory results. 
\end{itemize}

In summary, (i) FSCIL is a challenging task that can not be simply solved by finetune or joint training. (ii) Despite the advances achieved by recent methods, some methods need concession when balancing the average accuracy \texttt{Avg.} and the \texttt{PD} value. In contrast, our proposed KANet not only achieves the highest \texttt{Avg.} but also the smallest \texttt{PD} on CIFAR100, CUB200, and ImageNet-R.

\begin{figure*}
\centering
\includegraphics[width=1.8\columnwidth]{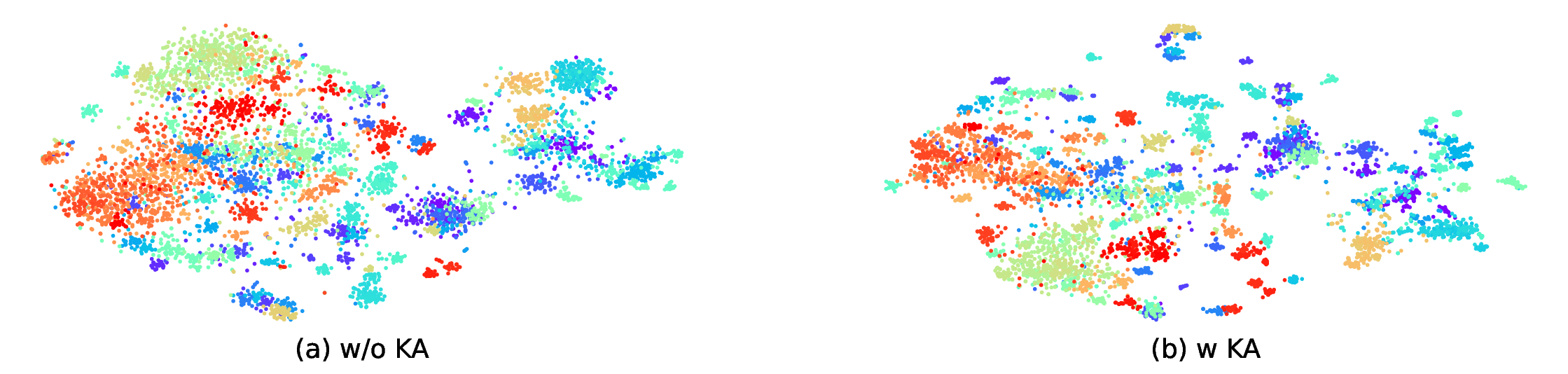}
\caption{t-SNE visualization on the resulting feature spaces generated by (a) removing and (b) using knowledge adapter (KA). 
\label{fig:tsne}}
\end{figure*}

\begin{table}\footnotesize
    \centering
     \caption{Ablation study on CUB200, where \textbf{KA} is the knowledge adapter, \textbf{IPEL} is the incremental pseudo episode learning, ``base'' and ``new'' represent the accuracy on base and new classes, respectively.}
    \setlength{\tabcolsep}{4.9mm}{
    \begin{tabular}{c|ccc}
    \toprule
    Equipped modules  & base$\uparrow$  & new$\uparrow$  & Avg.$\uparrow$   \\
    \midrule
    None                        & {71.26} & {44.73} & {63.98}  \\
    KA                          & {77.69} & {59.18} & {72.40}  \\
    \midrule
    KA \& IPEL                  & \textbf{78.42} & \textbf{60.22} & \textbf{73.17}\\
    \bottomrule
    \end{tabular}}
     \label{tab:ablat}
     
\end{table}

\subsection{Ablation study.}
To validate the respective contribution of knowledge adapter (KA) and incremental pseudo episode learning (IPEL), we conduct several ablation experiments on CUB200. Table~\ref{tab:ablat} (row 1) removes the KA and IPEL and uses the CLIP to perform FSCIL. The performance has a significant drop suggesting that directly using the CLIP can not solve the FSCIL well. Table~\ref{tab:ablat} (row 2) removes the IPEL and trains the KA using the standard training paradigm. Compared to the performance given by directly using the CLIP, the accuracy on base and new classes and the average accuracy have a significant improvement, demonstrating that the KA can adapt the CLIP to FSCIL well by fusing the general knowledge from the CLIP and expert knowledge corresponds to FSCIL. Table~\ref{tab:ablat} (row 3) uses the IPEL to train the KA and achieves the best performance on both base and new classes and the average accuracy. The results validate that the IPEL is effective.

To provide further insight into the KA, we plot the data embeddings of all classes in the low-dimension space with t-SNE\cite{van2008visualizing}. The results are shown in Figure~\ref{fig:tsne}, where the baseline is given by the CLIP. We can see that the proposed KA results in more clustered representations compared to the CLIP.

\begin{figure*}
\centering
\includegraphics[width=1.9\columnwidth]{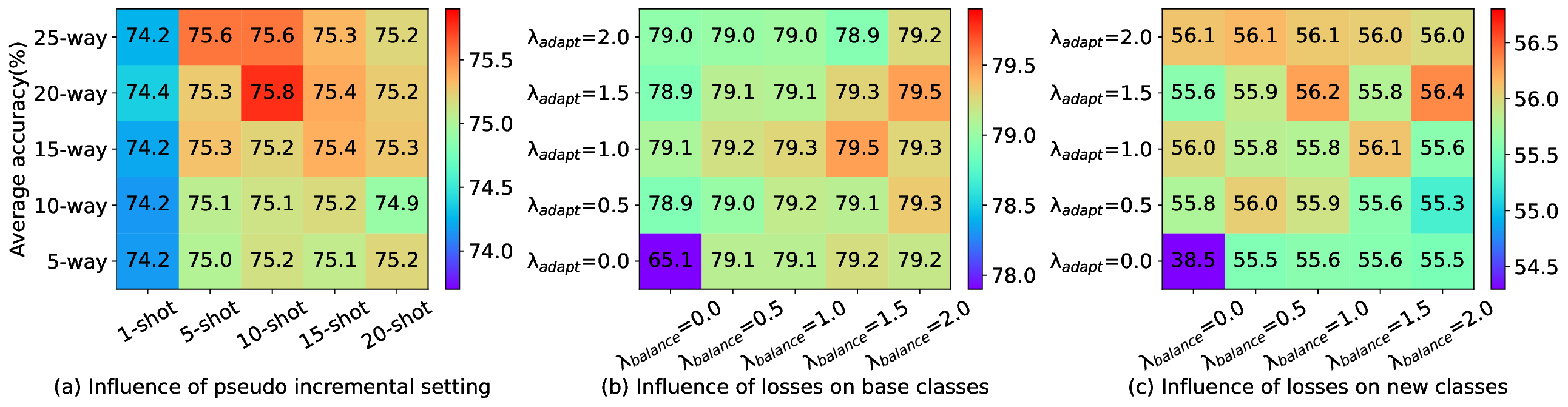}
\caption{Analysis of incremental pseudo episode learning under different conditions on CIFAR100. 
\label{fig:ipel}}
\end{figure*}

\subsection{Discussion}
\label{subsec:analy}

\begin{figure*}
\centering
\includegraphics[width=1.85\columnwidth]{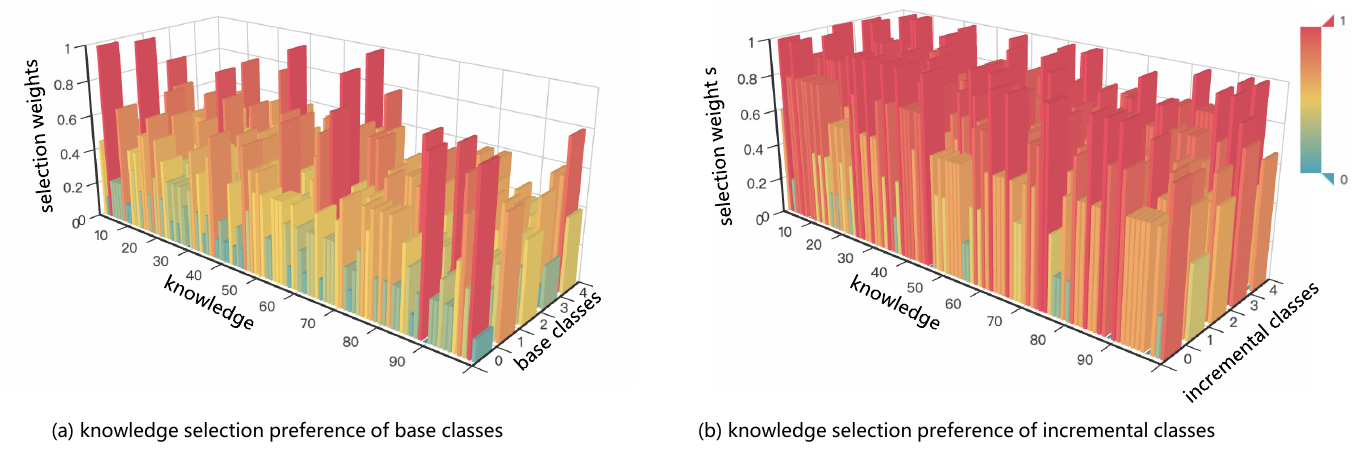}
\caption{Visualization of attention weights between different classes and the data-specific knowledge summarized in the knowledge vector library, where 5 base classes and 5 incremental classes are selected.
\label{fig:know_prefer}}
\end{figure*}

\subsubsection{Incremental pseudo episode learning.}
To analyze the influence of incremental pseudo episode learning, we first try different pseudo incremental settings and observe the corresponding influence on average accuracy. Then, we try different weights of $\mathcal{L}_{adapt}$ and $\mathcal{L}_{balance}$ and observe the corresponding influence on base and new classes.
\begin{itemize}
    \item As we can see from Figure~\ref{fig:ipel}(a), {a relatively larger number of pseudo new classes (way) and number of pseudo training samples(shot) can help our proposed method achieve better performance. Especially, when using a 20-way-10-shot setting to construct pseudo incremental task helps our proposed method get the highest average accuracy.}
    \item As can be observed in Figure~\ref{fig:ipel}(b), the accuracy on old classes given by using only the $\mathcal{L}_{adapt}$ is smaller than that given by using only the  $\mathcal{L}_{balance}$. This indicates that  $\mathcal{L}_{balance}$ contributes more to maintaining the performance of old classes compared to $\mathcal{L}_{adapt}$. 
    \item As can be observed in Figure~\ref{fig:ipel}(c), the accuracy on new classes given by using only the $\mathcal{L}_{adapt}$ is larger than that given by using only the  $\mathcal{L}_{balance}$. This indicates that  $\mathcal{L}_{adapt}$ contributes more to improving the performance on new classes compared to $\mathcal{L}_{balance}$. Particularly, setting the weight of $\mathcal{L}_{adapt}$ to 1.5 and the weight of $\mathcal{L}_{balance}$ to 2.0 helps our method achieve the highest accuracy on both the old and new classes.
\end{itemize}

\subsubsection{Data-specific knowledge}
To explore the effectiveness of the data-specific knowledge, one of the key components in our proposed method, we remove this component to observe the corresponding accuracy on CUB200 and CIFAR100. As we can see from table~\ref{tab:dsk}, removing the data-specific knowledge drops the performance on old classes indicating that this component helps improve the model's stability. Further, removing the data-specific knowledge drops the performance on new classes indicating that this component helps improve the model's plasticity. Specifically, on CIFAR100, removing this module leads to a larger performance degeneration on new classes compared that on old classes, this indicates that using the data-specific knowledge may prefer to improving the model's plasticity. In summary, using the data-specific knowledge not only improve the stability and plasticity of our proposed method.

\begin{table}[t]\footnotesize
    \centering

     \caption{The influence of data-specific knowledge $\mathcal{M}$ on CUB200 and CIFAR100, where ``base'' and ``new'' represent the accuracy on base and new classes, respectively.}
    \setlength{\tabcolsep}{2.1mm}{
    \begin{tabular}{c|ccc|ccc}
    \toprule
    \multirow{2}{*}{Method} & \multicolumn{3}{c|}{CUB200}  & \multicolumn{3}{c}{CIFAR100}\\
    \cmidrule{2-7}
    {}  & base$\uparrow$  & new$\uparrow$  & Avg.$\uparrow$  & base$\uparrow$  & new$\uparrow$  & Avg.$\uparrow$ \\
    \midrule
    w/o $\mathcal{M}$                       & {75.21} & {58.66} & {70.50} & {73.85} & {45.85} & {69.32} \\
    w $\mathcal{M}$                         & \textbf{78.42} & \textbf{60.22} & \textbf{73.17} & \textbf{79.47} & \textbf{56.35} & \textbf{75.77} \\
    \bottomrule
    \end{tabular}}
     \label{tab:dsk}
     
\end{table}

\subsubsection{Query-based knowledge fusion mechanism}
\label{layer_config}
Our proposed knowledge adapter fuses the general knowledge from the CLIP and data-specific knowledge by query-based knowledge fusion(QKF) mechanism.
Concretely, we adopt the cross-attention to achieve the QKF mechanism, which means that we will first compute the attention weights between the token and data-specific knowledge, and then use them to query potential corresponding items from the knowledge vector library $\mathcal{M}$ to adaptively enhance the token. To observe the behavior of QKF, we visualize the attention weights of different classes to the knowledge in $\mathcal{M}$. As can be observed in  Figure \ref{fig:know_prefer}, for incremental classes, we find that they prefer to select a large number of various knowledge to enhance themselves. In contrast, the number of knowledge selected by base classes is relatively smaller. We think the primary reason stems from that the token enhancement of incremental classes demands a larger amount of knowledge, {but the training samples is scarce.}

\begin{table}[t]\footnotesize
    \centering
        \caption{Comparison of different layer configurations on CIFAR100, where the \textbf{KS} represents the knowledge summary layer, \textbf{KF} represents the knowledge fusion layer.}
        \setlength{\tabcolsep}{2.3mm}{
        \begin{tabular}{c|cc|cc|cc}
        \toprule
        \multirow{2}{*}{\# KS} & \multicolumn{2}{c|}{base classes} & \multicolumn{2}{c|}{new classes} & \multicolumn{2}{c}{all classes} \\
        \cmidrule{2-7}
        {} & {\# KF} & {\texttt{Acc.}} & {\# KF} & {\texttt{Acc.}} & {\# KF} & {\texttt{Avg.}}\\
        \midrule
        {1} & {10} & {79.33}  & {10} & {56.08}  & {10} & {75.70} \\
        {2} & {10} & {79.32}  & {11} & {56.28} & {10} & {75.68} \\
        {3} & {10} & {79.25} & \textbf{11} & \textbf{56.67} & {10} & {75.61} \\
        {4} & \textbf{10} & \textbf{79.47} & {10} & {56.35} & \textbf{10} & \textbf{75.77} \\
        {5} & {7} & {79.15}  & {10} & {56.35} & {10} & {75.63} \\
        {6} & {10} & {79.28}  & {11} & {56.15} & {10} & {75.63} \\
        {7} & {9} & {79.20} & {10} & {56.35} & {10} & {75.58} \\
        {8} & {2} & {79.17} & {11} & {56.23} & {10} & {75.51} \\
        {9} & {7} & {79.35} & {11} & {56.23} & {10} & {75.56} \\
        {10} & {2} & {79.27} & {11} & {56.08} & {10} & {75.59} \\
        {11} & {7} & {79.22} & {11} & {56.05} & {10} & {75.43} \\
        {12} & {10} & {79.40} & {10} & {56.15} & {10} & {75.68} \\
    
        \bottomrule 
        \end{tabular}}
        \label{tab:mem_upd}
\end{table}

\subsubsection{Layer configurations}
\label{subsubsec:layer_conf}
In our default setting, we summarize the data-specific knowledge from the forth encoder layer, and plug the knowledge adapter into the tenth encoder layer. This subsection studies the effect of different configurations of these two types of layers, where we use ``KS'' (Knowledge Summary) to denote the layer used to summarize the data-specific knowledge, ``KF'' (Knowledge Fusion) to denote the layer for plugging the knowledge adapter. As can be observed in Table~\ref{tab:mem_upd}, though satisfactory performance can be achieved by selecting an optimal KF layer for each KS layer, our proposed method generally prefers a relatively shallower KS layer and a relatively deeper KF layer. Specifically, based on the average accuracy, setting the KS layer to 4 and the KF layer to 10 yields the best performance.

\subsubsection{Performance under different backbones} 
To {investigate} the effectiveness of our proposed method under different backbones, we conduct several comparative experiments using various backbones. The results are shown in Table~\ref{tab:backbone}, where the baseline refers to directly using the pretrained model to perform FSCIL. The significant improvements across different backbones demonstrate that our proposed method is capable of effectively adapting the ViT-based pretrained model to FSCIL.

\begin{figure*}
\centering
\includegraphics[width=2.0\columnwidth]{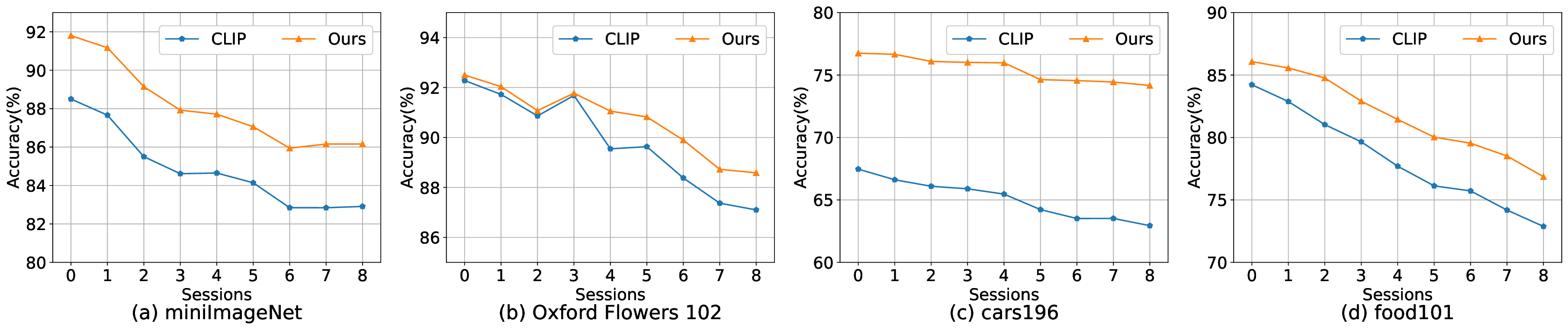}
\caption{The performance of our proposed method and CLIP on different datasets. Our proposed method achieves substantial improvement over CLIP.
\label{fig:inc_other_dataset}}
\end{figure*}

\begin{table}[h]\footnotesize
    \centering

    \caption{Performance comparisons under different backbones on CIFAR100.} 
    \setlength{\tabcolsep}{2.7mm}{
    \begin{tabular}{l|c|ccc}
    \toprule
    {Method} & {Backbone} & base$\uparrow$  & new$\uparrow$  & Avg.$\uparrow$ \\
    \midrule
    Baseline \cite{radford2021learning} &  \multirow{2}{*}{CLIP-ViT-B/32}               & {71.26} & {44.73} & {63.98} \\
    KANet(Ours)                         & {}                                            & \textbf{78.42} & \textbf{60.22} & \textbf{73.17} \\
    \midrule
    Baseline \cite{radford2021learning} & \multirow{2}{*}{CLIP-ViT-B/16} & {70.12} & {42.38} & {64.93} \\
    KANet(Ours)                         & {}                             & \textbf{81.83} & \textbf{60.33} & \textbf{78.17} \\
    \midrule
    Baseline \cite{dosovitskiy2020image} &  \multirow{2}{*}{ViT-B/16}      & {77.05} & {60.13} & {74.17}  \\
    KANet(Ours)                         & {}                               & \textbf{87.55} & \textbf{76.18} & \textbf{85.87} \\
    
    \bottomrule 
    \end{tabular}}
    \label{tab:backbone}
\end{table}

\begin{table}[t]\small
    \centering

     \caption{The few-shot class-incremental setting of different datasets, where $N_{base}$, $N_{new}$, and $N_{sess.}$ refer to the number of base classes, new classes, and sessions, respectively.}
    \setlength{\tabcolsep}{4.3mm}{
    \begin{tabular}{c|ccc}
    \toprule
    dataset  & $N_{base}$  & $N_{new}$  & $N_{sess.}$   \\
    \midrule
    \textit{mini}ImageNet       & 60 & 40 & 1+8  \\
    Oxford Flowers 102          & 62 & 40 & 1+8  \\
    cars196                     & 156 & 40 & 1+8 \\
    food101                     & 61 & 40 & 1+8  \\
    \bottomrule
    \end{tabular}}
     \label{tab:dataset}
     
\end{table}

\subsection{More validations for our proposed method}
To further {explore} the effectiveness of our method, we {construct more few-shot class-incremental learning datasets and evaluate our model}, including \textit{mini}ImageNet which is the subset of ImageNet~\cite{2015ImageNet}, the fine-grained datasets Oxford Flower102~\cite{nilsback2008automated}, cars196~\cite{KrauseStarkDengFei-Fei_3DRR2013}, and food101~\cite{bossard14}. We use the setting shown in Table ~\ref{tab:dataset} to split each dataset and report the experimental results in Figure~\ref{fig:inc_other_dataset}. We can see that despite varying degrees of data leakage across datasets, our proposed method consistently outperforms CLIP.

\section{Conclusion}\label{sec:conclusion and future work}

In this paper, we explore the adaptation of CLIP and its efficacy in few-shot class-incremental learning.
Regarding the two challenges that exist in adapting the CLIP to match the context of FSCIL, we propose the knowledge adapter (KA) and the incremental pseudo episode training (IPEL), resulting in the Knowledge Adapter Network (KANet). 
The KA fuses the knowledge from the CLIP, data-specific knowledge extracted from the data, and the task-dependent learned from the base session to harvest a more comprehensive representation to benefit the few-shot class-incremental learning tasks.
The IPEL mimics the real incremental setting and constructs a series of pseudo learning tasks with the data sampled from the base session to transfer the knowledge learned from the base session to the incremental sessions.
The experimental results on CIFAR100, CUB200, and ImageNet-R demonstrate the proposed KANet is an effective approach for FSCIL.

\section*{Acknowledgements}
This work was supported by the NSFC under Grant 62272380 and 62103317.

{\small
\bibliographystyle{ieee_fullname}
\bibliography{ref}
}
\clearpage

\appendix




\end{document}